\documentclass[a4paper]{article}

\usepackage[english]{babel}
\usepackage[utf8]{inputenc}
\usepackage{amsmath}
\usepackage{graphicx}
\usepackage[colorinlistoftodos]{todonotes}

\title{Data Augmentation for Modeling Human Personality: The Dexter Machine}

\author{Yair Neuman, Vladyslav~Kozhukhov and Dan Vilenchik}

\date{\today}

\begin{document}
\maketitle

\begin{abstract}
Modeling human personality is important for several AI challenges, from the engineering of artificial psychotherapists to the design of persona bots. However, the field of computational personality analysis heavily relies on labeled data, which may be expensive, difficult or impossible to get. This problem is amplified when dealing with rare personality types or disorders (e.g., the anti-social psychopathic personality disorder). In this context, we developed a text-based data augmentation approach for human personality (PEDANT). PEDANT doesn't rely on the common type of labeled data but on the generative pre-trained model (GPT) combined with domain expertise. Testing the methodology on three different datasets, provides results that support the quality of the generated data.  \end{abstract}

\section{Introduction}\label{sec:introduction}

Personality concerns the individual's relatively
stable pattern of thoughts, emotions and behaviors \cite{CorrMath}. There are various personality theories from the Big Five \cite{goldberg1990alternative}  to Affective Neuroscience \cite{davis2011brain} and Mischel's contextual approach to personality \cite{mischel2004toward}. In this paper, we adhere to the clinical approach represented by the Psychodynamic Diagnostic Manual (PDM) \cite{lingiardi2017psychodynamic} and SWAP \cite{shedler2007shedler}, which is highly relevant for diagnosis, and research \cite{lingiardi2015psychodynamic}. 
According to the PDM approach, personality types are stable configurations characterized by key features such as the individual's core beliefs about self and others. For instance, a depressive personality is characterized by self-criticism and accompanied by the belief that "something is essentially bad about me". 

Current computational personality research is almost exclusively focused on features'-based data-driven classification involving the prediction of a personality class label. Accomplishing such tasks  relies on the availability of a large amount of high-quality {\em labeled} data (e.g.,~\cite{neuman2016computational,losada2016test,bachrach2012personality}). However, obtaining such data may be expensive, difficult, or impossible for various reasons. For instance, the prevalence of the anti-social psychopathic personality disorder in the population is low ~\cite{sanz2021prevalence,werner2015epidemiology,holzer2020prevalence}, and it is currently impossible to gain access to a massive dataset of labeled texts produced by clinically diagnosed psychopaths. High-quality diagnostic procedures, such as SWAP, are costly as they require human expertise and significant time to complete. While self-reported questionnaires for personality assessment are available, they rely on the collaboration of the diagnosed individual and their ability to provide a valid self-assessment, which in the case of the anti-social personality disorder, for instance, is not trivial to gain.     

In the face of these challenges, a natural solution for data scarcity is data augmentation, intensively developed in computer vision but "relatively under-explored" in NLP, where the generation of effective augmented examples is "less obvious" \cite{feng2021survey,dao2019kernel}. To illustrate the challenges in textual data augmentation, we ran the SOTA data-augmentation pipeline LAMBADA \cite{anaby2020not} to generate 200 sentences out of a seed set of 20 sentences (see appendix) expressing a clear psychopathic signature. This attempt to produce artificial "psychopathic" sentences resulted in only 100 unique sentences, where the vast majority of sentences were either one of the seed sentences or a simple paraphrasing thereof.

Data augmentation is typically viewed as the process of increasing the amount of data by adding slightly modified copies of \emph{already existing} labeled data. In some cases, there is no labeled data at all or a very small quantity which precludes proper augmentation (as our experiment with LAMBADA suggests). In this paper, we offer a solution to these cases by using unlabelled data and adding domain expert input to compensate for the absence of labeled data (once can view labeled data as domain expert knowledge). 

A constructive approach to personality modeling may be found in the revolutionary large language models recently introduced to NLP (e.g.,
GPT-2) \cite{radford2018improving,radford2019language}. Recently, \cite{wolf2019transfertransfo} and \cite{golovanov2019large} showed that the GPT model, once fine-tuned, can be useful in the domain of personal conversations. Their approach led to substantial improvements in the PersonaChat data set, showcasing the potential of exploiting large pre-trained generative models in the conversational domain. However, these advancements do not naively imply anything for modeling personality types, as the poor results obtained from LAMBADA, which is based on GPT technology, show. Indeed, {\em personalized} chit-chat models, \cite{zhang2018personalizing} use the notion of {\em personalization}~(e.g., age) which is different from the psychodynamic approach used in this paper. 

\subsection{\bf Our contribution.}

We present a novel  personality data augmentation approach, PEDANT (PErsonality Data AugmeNTation), using (1) a generative pre-trained model (GPT) combined with (2) domain expertise (the domain expert is the first author who has intensively studied and published about personality) while relying only on (3) unlabeled text. 

PEDANT operates in two phases. In the first phase, unlabeled data relevant to the selected personality type is harvested from online resources; this data is then used to train a generative language model. In the second phase, the language model is repeatedly prompted to complete a set of seed sentences carefully crafted by the domain expert. All these completions are then filtered and ranked according to a scoring function that the domain expert pre-defined; the top $k$ sentences are the output of PEDANT. 



We implement PEDANT with regard to a specific personality type: the anti-social psychopathic personality \cite{shedler2007shedler}; we call this particular pipeline Dexter. This type of personality is suitable for validating our approach as the prevalence of psychopathic personality disorder is extremely low, and a labeled corpus of naturally produced texts of diagnosed psychopaths does not exist. The texts that we harvest for the first phase of PEDANT come from a few fictive characters from the cinema and TV (e.g., Dexter the psychopath from the TV series "Dexter") and from Reddit forums such as \texttt{r/psychopath}. The second phase, where domain expertise is used, is described in detail in Sections \ref{sec:seed} and \ref{sec:method:ranking}.

We validated Dexter using a downstream text classification task, as common in other works that deal with the evaluation of  data augmentation pipelines \cite{ruder2018strong,anaby2020not}.
We used the data generated by Dexter to train a classifier and then tested it on three offensive-speech datasets that cover different dimensions of the psychopathic personality (e.g., lack of empathy, toxicity, and being manipulative). 

For comparison, we tested two SOTA data augmentation pipelines, LAMBADA \cite{anaby2020not}, and LeakGAN \cite{LeakGan}. The Dexter dataset produced a classifier that ranked first (by a large gap) in all three datasets.
The complete detail of both experiments appears in Section \ref{sec:results}. 

\subsection{Illustration}\label{sec:illustration}
It is a non-trivial task to evaluate the extent to which the resulting generative model (the outcome of Dexter) reflects the psychopathic mind. One way mentioned above is via a downstream task that uses data generated by the model. Another way is to have a host of personality domain experts chat with the model. While we did not have the resources to perform this expensive and laborious task, we invite the reader to peek into such a possible Q\&A session and to judge for herself the change in personality.  

Below is a comparison of the output of a GPT model before and after fine-tuning on the harvested psychopathic-related texts. In the examples below, GPT is prompt with psychopathic statements chosen by the domain expert, and its completion is recorded; a couple more examples are given in the appendix.

\vspace{1.2mm}

\noindent Statement: {\it I act impulsively, without regard for the consequences of my actions.} \\
\hfill GPT-2 before fine-tuning:  {\it I know when someone is suffering and I feel bad}.\\
\hfill GPT-3 davinci:  {\it I have difficulty organizing tasks and activities. I have difficulty with time management. }.\\
\hfill GPT-3 curie:  {\it My friends are more important to me than my family. I dress in a way that is not in keeping with the standards of my community}.\\
\hfill chatGPT:  {\it
It's natural to act impulsively at times, but if you find that you frequently act without considering the consequences of your actions, it may be helpful to try to develop more self-control.}
\hfill {\bf GPT-2 after fine-tuning:} {\it I have an abundance of rage, and I can turn to it with no consideration for consequences.}
\vspace{1.2mm}

\noindent Statement: {\it I feel like a hunter seeking a prey.} \\
\hfill GPT-2 before fine-tuning:  {\it I don’t want to be a bad person}.\\
\hfill GPT-3 davinci:  {\it I want to let my heart feel.}\\
\hfill GPT-3 curie:  {\it I'm lost in this forest ! I'll be crushed by this forest! }\\
\hfill chatGPT:  {\it It's important to remember that other people are not prey to be hunted. It's never okay to view or treat others as if they are objects or targets to be pursued or exploited.}\\
\hfill {\bf GPT-2 after fine-tuning:} {\it I don’t care
about it, I just want to kill and eat.}

\section{Related work}
Data augmentation methods have been explored to address the imbalance of datasets challenge in supervised classification tasks. Noise injection or attribute modification techniques were commonly applied to generate synthetic data for image and sound classification tasks \cite{shorten2019survey,tran2017bayesian,salamon2017deep}. However,  such techniques do not extend to text due to the categorical nature of words and the sequential nature of text.

We locate our work in the context of text data augmentation, \cite{feng2021survey,shorten2021text,anaby2020not}. ``Previous-generation" textual data augmentation approaches focus on sample alteration \cite{kobayashi2018contextual,mueller2016siamese,wu2019conditional,rizos2019augment,ibrahim2018imbalanced}, in which a single sentence is altered to generate a new sentence
primarily by synonyms replacements. Presumably, methods that make only local changes will produce sentences with a structure similar to the original ones, thus yielding low corpus-level variability. 

Other recent approaches to textual data augmentation that generate whole sentences rather than making a few local changes include using variational autoencoding \cite{kingma2013auto}, paraphrasing \cite{kumar2019submodular} and methods based on generative adversarial networks \cite{tanaka2019data,LeakGan,cao-lee-2020-hategan}. 

Recent progress in NLP has been marked by the emergence of large language models (i.e., transformers) such as GPT-2 \cite{radford2019language}. GPT-based language models scored high in open-domain dialogue generation tasks \cite{golovanov2019large,wolf2019transfertransfo,zhang2019dialogpt}. The data-augmentation pipeline presented in \cite{anaby2020not} uses GPT technology to generate themed synthetic text. The idea behind \cite{anaby2020not} involves fine-tuning a GPT model to a specific task using existing labeled data. Using the fine-tuned model and given a class label, new sentences for the class are generated. The sentences are filtered with a  classifier trained on the original data. 

While our pipeline is similar to \cite{anaby2020not} in flavor (a fine-tuning step followed by a filtering step), it is different in two key aspects. We use unlabelled data for the fine-tuning step. This allows us to fine-tune the GPT model with a large amount of, possibly slightly lower quality, data. \cite{anaby2020not} use labeled data for the fine-tuning step; thus the quality of the augmentation depends on the amount of available text. Second, our filtering is also done in an unsupervised manner, replacing the need for labeled data for training a classifier with the knowledge of a domain expert. These two key differences make our pipeline useful for data generation for rare classes, such as rare personality types, where labeled data is scarce or non-existent. Indeed,  comparing the performance of \cite{anaby2020not} to Dexter corroborates the latter.

\begin{figure*}[ht!]
\centering
\includegraphics[width=1\linewidth]{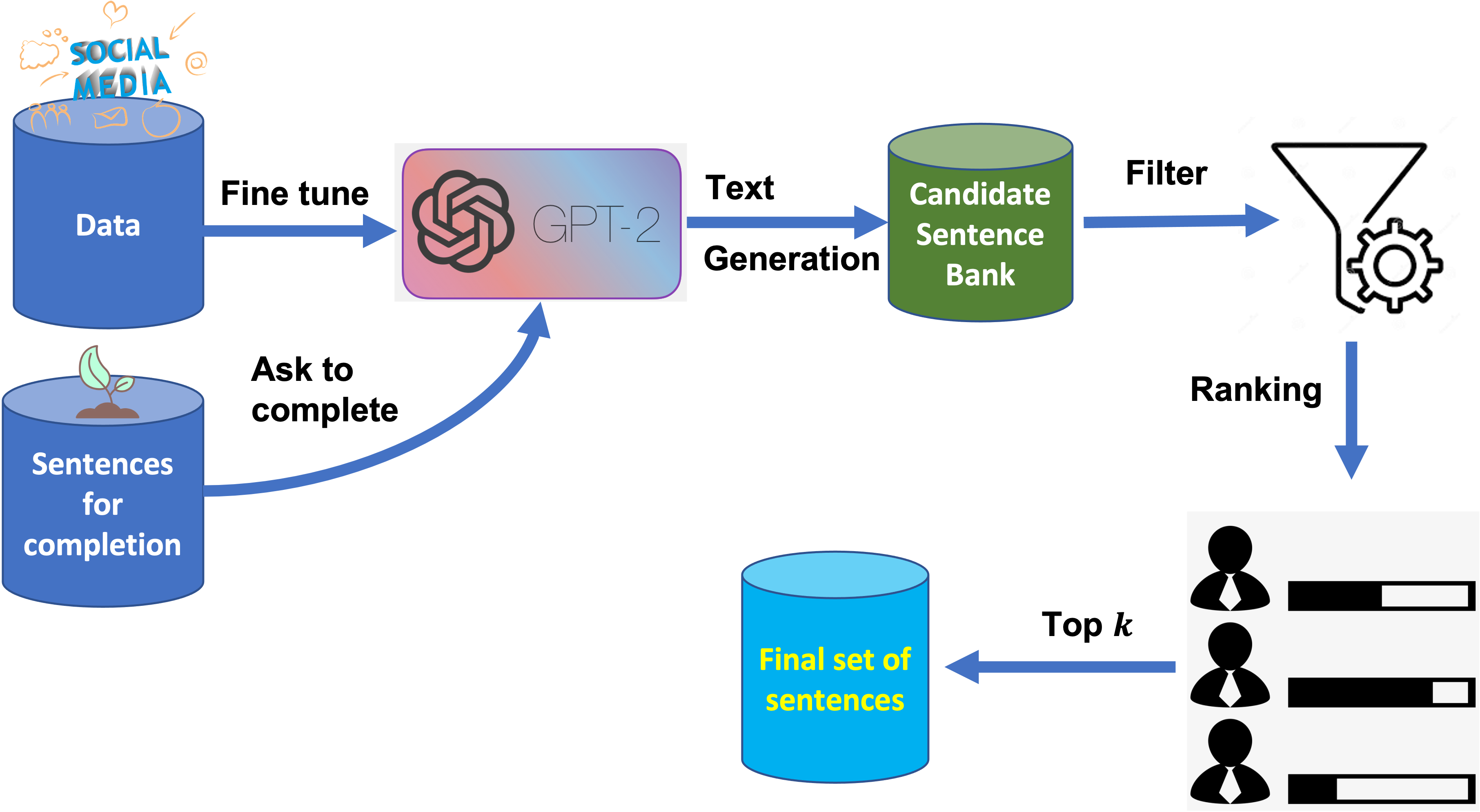}
\caption{Pipeline for PEDANT. GPT-2 is trained and prompted to complete a carefully chosen seed of sentences. The completions are filtered and ranked using similarity to relevant words chosen using domain expertise.}
\label{fig:pipeline}
\end{figure*}

\section{Methodology}\label{sec:method}
The pipeline for generating  data for a given personality type, illustrated in Figure \ref{fig:pipeline}, is composed of the following stages (the actual parameter values that we've used are given in this general description):
\begin{enumerate}
    \item Texts produced by a few fictive characters (e.g., Dexter) and secondary sources (e.g., Reddit forums discussing the personality style) are collected to form a preliminary dataset. Let $\cal{D}$ be that preliminary dataset.
    \item A pre-trained language model is fine-tuned on $\cal{D}$. Let $\cal{G}$ be the obtained model
    \item A domain-expert hand-crafted set of $s=40$ seed sentences
    representing the personality's beliefs about self/others is prepared; $\cal{G}$ is prompted to complete each seed $c=200$ times, for  a total of $n=s\cdot c=8000$ candidate sentences.
    \item Based on domain expertise, a hand-crafted vector $\cal{F}$, containing $f$ words that are typical of the personality type, is assembled. The $n$ candidate sentences are being filtered and ranked according to their cosine similarity with $\cal{F}$.
    \item The top $k=2000$ sentences compose the output.
\end{enumerate}
We now describe how we customized this general pipeline to the psychopathic personality. 

\subsection{The preliminary datasets}\label{sec:traindata}
Following \cite{serban2015survey,coppersmith2014quantifying}
we used data from movie scripts -- the text produced by three well-known fictive psychopathic characters: The Joker in the movie "The Joker", Bateman in the movie "American Psycho" and Dexter from the TV series "Dexter". In addition, we collected all texts from Reddit discussion groups dealing with psychopathy (\texttt{r/psychopath, r/sociopath, r/antisocial}). 

After cleaning the data by applying a spell checker \cite{AutoCorrect}, removing emojis, duplicates, hyperlinks, and spam messages \cite{redditcleaner}, the preliminary cleaned dataset consisted of 1,320,552 tokens. 

\subsection{The sentences completion seed set}\label{sec:seed}
Our domain expert manually prepared 20 seed sentences representing the psychopath's "beliefs about self" (e.g., "I take advantage of others whenever I can") and 20 seed sentences representing the psychopath's ''beliefs about others" (e.g., "Human beings are weak"). The complete list of seed sentences appears in the appendix. 

The number '20' is somewhat arbitrary. While testing with other seed sizes, we found that 20 was the minimal number that gave good results, considering the computational constraints such as space and running time.

\subsection{Training GPT-2 and generating sentences}

Our starting point is the pre-trained GPT-2 with 1.5B parameters accessed via the popular HuggingFace API \cite{wolf2019huggingface}. We chose GPT-2 as it is currently one of the most useful language generation model. (The newer GPT-3 is still not open source, and it's harder to work with, e.g., fine-tuning on a large text like our preliminary dataset). Next, we fine-tuned the pre-trained GPT-2 model on the entire preliminary dataset (the fictive characters text and the Reddit data) using the task of predicting the next word of the sentence~\cite{radford2019language}. The parameters we used were: learning rate=0.0001, model name=`1558M', batch size=4, optimizer = `adafactor', steps = 10000 and the cross entropy loss function.

We prompted the fine-tuned GPT-2 model on each of the 40 seed sentences ( see the appendix for the full list), producing 200 sentence completions for each sentence. Using the experts' judgment of two psychologists, we qualitatively evaluated a random sample of these sentences. We concluded that the best completions were obtained with the following parameters: length=50,temperature=0.7, top $k$ = 50, top $p =0.90$.

We used the free Google Colab resources with Tesla P100 GPUs for this part.

\subsection{Filtering and Ranking}\label{sec:method:ranking}
We applied filtering and ranking to the 8000 sentence completions. First, we removed sentences that (1) include the trivial words: psychopath, antisocial, and sociopath; (2) are duplicates of other sentences;
(3) contain less than three words;
(4) end with a stop word;
    (5) are emotionally neutral or have a higher positive than a negative sentiment (we used NLTK to estimate the sentiment);
    (6) are simple paraphrases of each other (via \cite{reimers2019sentence}).

For the ranking task, we identified words significantly collocated with the target word "psychopath" in the iWeb repository \cite{davies2019advantages}. The domain expert selected 28 to form a "psychopathic vector" (see the appendix). Next, we used the vectorial semantic approach for personality assessment \cite{neuman2014vectorial} and measured the cosine similarity between each filtered candidate sentence completion and the psychopathy vector. 
For each of the 40 seed sentences, we selected the 50 completions that scored highest on the cosine similarity test. The output was a set of 1735 synthetically generated sentences that are supposed to represent a psychopathic mind (if for some seed less than 50 completions passed the filtering step, we took all of them). 

\section{Data}\label{sec:data}
Our setting inherently precludes the existence of a large labeled bench-marking dataset of text written by clinically diagnosed psychopaths. However, as the antisocial psychopathic personality is composed of several dimensions (e.g., lack of empathy), we tested our approach on labeled datasets hypothesized to share one dimension or more with this personality type.


\subsection{Test Data}\label{sec:test_data}
We now describe the three datasets that we've used to evaluate the performance of Dexter and two other data augmentation pipelines.

\vspace{1.2mm}

\noindent{\bf{Sexual predators.}}
Sexual predators share with psychopaths at least two psychological dimensions: being manipulative and lacking empathy, as indicated by the correlation between sexual offending and psychopathy \cite{gretton2001psychopathy,krstic2018using,sohn2020utility}. Our first dataset  was a labeled data set of texts produced by 142 sexual predators and 97,689 non-predators \cite{inches2012overview}. 

\vspace{1.2mm}

\noindent{\bf{Empathy.}}
Psychopaths are characterized by a lack of empathy. Our second data set consists of interactions between help seekers in an online call center for mental health support \cite{sharma2020computational}. Labeled texts of the mental health supporters (responders) are provided. Responders are tagged according to three increasing levels of empathy: “0” ($N=2037$), “1” ($N=895$), and “2” ($N =152$).  
Unlike the other two datasets, the empathy dataset does not contain a natural positive class, as an empathy score of 0 does not necessarily imply a strong negative personality. 

\vspace{1.2mm}

\noindent{\bf Cyberbullying.} Cyberbullying may have a clear psychopathic signature, given the reported association between the psychopathic mind and sadism \cite{meere2017everyday,sest2017constructing}. We have used the labeled toxic-text subset of the cyberbullying dataset \cite{Cyberbullying} that contains 12,168 toxic vs. 14,874 non-toxic texts. Unlike the previous two datasets, this one is labeled at the message level. Each message consists of several sentences, and the entire message is assigned the label ``toxic" if there are ``enough" toxic sentences (the exact labeling procedure is described in the original paper \cite{Cyberbullying}). 

\subsection{Train Data}\label{sec:train_data}
We now describe the datasets we used to fine-tune the BERT-base-uncased model \cite{devlin2018bert}, which we then ran to classify the aforementioned test datasets. The data statistics is summarized in Table \ref{tab:data}.

\vspace{1.2mm}

\noindent{\bf The Dexter dataset.} This dataset contains 3400 sentences; 1700 sentences are the output of Dexter, which serve as the positive class, and 1700 sentences from various Reddit discussion groups that serve as the negative class. These sentences were selected by first sampling 8000 random sentences from various Reddit groups, then filtering and cleaning them according to the same procedure that was applied to the psychopathic texts. Finally, the 1700 sentences with the lowest psychopathic score were chosen.

\vspace{1.2mm}

\noindent{\bf The Dexter-minus and the PRELIM dataset}. These two datasets allow us to evaluate the importance of the different stages in the Dexter pipeline (Figure \ref{fig:pipeline}). The Dexter-minus dataset follows the same pipeline as Dexter just that the fine-tuning of the GPT is skipped. The PERLIM dataset shortcuts the GPT step altogether and proceeds directly to the filtering and ranking step.

\vspace{1.2mm}
\begin{table} [t!]
\small
\centering
\begin{tabular}{ |c| c| c| }
\hline
  Dataset & \#Pos. & \#Neg   \\ \hline
  
{\bf Sexual Predators }& 142 & 97,689  \\ \hline
  {\bf Empathy}  & 2,037 & 152  \\ \hline
  {\bf Cyberbullying} & 12,168 & 14,784  \\ \hline 
\end{tabular}\caption{Summary statistics for the three datasets described in Section \ref{sec:data}. The number of samples in the positive and negative class are shown.}
\label{tab:data}
\end{table} 

To compare the performance of Dexter against the two SOTA data-augmentation pipelines, we created the following two synthetic datasets.

\vspace{1.2mm}

\noindent{\bf The LAMBADA dataset.}
To train LAMBADA, we used the Papers-With-Code recommended implementation of LAMBADA \cite{ma2019nlpaug}. We augmented GPT-2 with two new classes, "\#beliefs about others" and "\# beliefs about self". Each class was seeded with the 20  sentences that our domain expert crafted (Section \ref{sec:seed}). All the training parameters and the code are available at \cite{ma2019nlpaug}. 

We then used the LAMBADA pipeline to generate 17,000 sentences, out of which we chose the best 1700 (this was the recommendation of the authors \cite{anaby2020not}, to generate 10x more sentences than needed). Specifically, we generated 8500 from the ``\#beliefs about others" class 
and 8500 from ``\#beliefs about self" (we used the following parameters for GPT-2: max length=50, top $k=10$, $p=0.85$). We ranked the sentences the same way we ranked ours: using cosine similarity to the psychopathic vector of words (Section \ref{sec:method:ranking}).

The negative class of the LAMBADA dataset is the same as Dexter's.

\vspace{1.2mm}

\noindent{\bf The LeakGAN dataset.} We trained LeakGAN using the official LeakGAN implementation \cite{LeakGit}.  LeakGAN is trained with the target class text. We used all the text in the preliminary dataset to that end (Section \ref{sec:traindata}) and the default parameters from the official implementation. We generated 1700 sentences using LeakGAN to serve as the positive class, and the negative class was the same as Dexter's.

\vspace{1.2mm}

\noindent{\bf External competition datasets.} This collection contains three  gold-standard competition offensive speech datasets: OffenseEval \cite{zampieri2019semeval}, HatEval \cite{basile2019semeval}, and AbuseEval \cite{caselli2020feel}. Each dataset contains roughly 10,000 labeled texts. We call the union of all three datasets the Golden dataset.


\section{Evaluation}\label{sec:eval}
To evaluate Dexter we've created a family of models using the aforementioned training datasets of Section \ref{sec:train_data}. Each model is named X@BERT, which means that the BERT-base-uncased model \cite{devlin2018bert} was fine-tuned using dataset X. If a '+' is appended, X@BERT+, this means that a preceding step of fine-tuning on OffenseEval \cite{zampieri2019semeval} was taking place.

\subsection{Evaluation procedure}\label{sec:eval_proced}
All the test datasets mentioned in Section \ref{sec:test_data}, Table \ref{tab:data}, are imbalanced to different degrees (the sexual predators dataset contained merely 0.001\% sexual predators). To allow comparison across the three datasets while avoiding misleading artifacts that such imbalanced data introduces, we down-sampled the majority class to obtain balanced sets.

The output of a BERT model is a number in $[0,1]$, the result of the last layer activation unit (soft-max in our case). This number may be thought of as the probability that BERT assigns the instance to belong to the positive class (in our case, "psychopath"). 
We define the {\em PsychoScore} of a user as the average output of the model over all the sentences produced by that user (each sentence is scored separately by the model). It is common practice to feed the BERT score into a simple classifier, like SVM, to find the optimal cut-off for the binary classification task \cite{raj2019recurrent}.

To evaluate the model on each test data set, we computed the 5-fold cross-validation F1 and Macro F1 scores. Each fold consisted of $n=100$ randomly sampled instances from each class, and split into 80\% train and 20\% test. We trained a soft-margin kernel SVM (we used the default Python sklearn module parameters $C=1$, kernel = RBF)  on the users' PsychoScores and  the corresponding label.

\begin{table} [t!]
\small
\centering
\begin{tabular}{ |c| c| c|c|c| }
\hline
  Model & Pred. & Emp. & Cyber. & Avg Rank\\ \hline
  
{\bf Dexter@BERT+ }& 1 & 1 & 2 & 1.33\\ \hline
  OffenseEval@BERT & 3 & 2 & 1 & 2\\ \hline
  {\bf Dexter@BERT} & 2 & 4 & 5 & 3.66\\ \hline
    HateEval@BERT & 6 & 3 & 6 & 5\\ \hline
  {\bf Dexter-@BERT } & 4 & 5 & 7 & 5.33\\ \hline
  Golden@BERT & 7 & 7  & 3 & 5.66\\ \hline
  AbuseEval@BERT & 7 & 7 & 4 & 6\\ \hline
   PRELIM@BERT & 8 & 8 & 8 & 8\\ \hline
  
\end{tabular}\caption{Summary of Table \ref{table:ClassificationResults}. The Dexter@BERT variants are  in bold.}
\label{tab:ranking}
\end{table}

\begin{table} [t!]
\small
\centering
\begin{tabular}{ |c| c| c|c|c| }
\hline
  Model & Pred. & Emp. & Cyber. & Avg Rank\\ \hline
  
{\bf Dexter@BERT+ }& 1 & 1 & 1 & 1\\ \hline
  {\bf Dexter@BERT} & 2 & 2 & 3 & 2.33\\ \hline
  LAMBADA@BERT+ & 4 & 4  & 2 & 3.33\\ \hline
  {\bf Dexter-@BERT } & 3 & 3 & 5 & 3.66\\ \hline 
  LeakGAN@BERT+ & 5 & 5 & 4 & 4.66\\ \hline
    LAMBADA@BERT & 7 & 6 & 6 & 6.33\\ \hline
  LeakGAN@BERT & 6 & 7 & 7 & 6.66\\ \hline

\end{tabular}\caption{Summary of Table \ref{table:ClassificationResults2}. The Dexter@BERT variants are  in bold.}
\label{tab:ranking2}
\end{table}

\subsection{Results}\label{sec:results}
The results of running the X@BERT models on the test datasets of Section \ref{sec:test_data} are summarized in Tables \ref{table:ClassificationResults} and \ref{table:ClassificationResults2}. Table \ref{table:ClassificationResults} reprots the comparison against the pre-trained offensive speech models, while Table \ref{table:ClassificationResults2} reports the comparison against LeakGAN and LAMBADA.
Table \ref{tab:ranking} summarizes Table \ref{table:ClassificationResults} with the overall average ranking across the three datasets and similarly Table \ref{tab:ranking2} summarizes Table \ref{table:ClassificationResults2}. Both show that the model Dexter@BERT+ 
ranked first, and Dexter@BERT came second (Table \ref{table:ClassificationResults2}) and third (Table \ref{table:ClassificationResults}) . The following key conclusions are read from the tables:

\begin{table*} [ht!]

\small
\label{table:results-4forum-posts}
\centering
\begin{tabular}{ |c| c | c c c c | }
\hline
\textbf{Data set}& \textbf{Model} & Precision & Recall &  F1 score &  Macro F1 score   \\

 \hline
& { Dexter@BERT+ }& $0.92$ & ${0.87}$ & $ 0.89$ & ${0.91} \pm {0.029}$ \\
& Dexter@BERT & 0.91 & 0.86 & 0.88 & $0.90 \pm 0.037$ \\
Sexual 
Predator Identification & OffenseEval@BERT  & 0.89 & 0.87 & 0.88 & $0.90 \pm 0.035$  \\
Competition \cite{inches2012overview}& Dexter-@BERT & 0.80 & 0.93 & 0.86 & $0.88 \pm 0.043$  \\

& HateEval@BERT & 0.95 & 0.5 & 0.65 & $0.75 \pm 0.011$\\
& Golden@BERT & 0.88	& 0.50 & 0.63 & $0.69 \pm 0.095$\\
& AbuseEval@BERT  & 0.73 & 0.58 & 0.51 & $0.53 \pm 0.133$ \\
& PRELIM@BERT & 0.51 & 1.00 & 0.68 & $0.38 \pm 0.024$ \\

\hline

&  Dexter@BERT+ & 0.66 & 0.80 & 0.72 & ${0.70} \pm 0.083$ \\
& OffenseEval@BERT & 0.61 & 0.81 & 0.69 & $0.64 \pm 0.081$  \\

 Empathy \cite{sharma2020computational}   &  HateEval@BERT  & 0.59 & 0.65 & 0.61 & $0.59 \pm 0.063$ \\
 & Dexter@BERT & 0.55 & 0.88 & 0.67 & $0.54 \pm 0.077$ \\

& Dexter-@BERT & 0.51 & 0.65 & 0.57 & $0.52 \pm 0.075$ \\

& Golden@BERT &0.42 & 0.93 & 0.58 & $0.38  \pm 0.061$ \\
& AbuseEval@BERT  & 0.44 & 0.85 & 0.58 & $0.37 \pm 0.018$  \\
&  PRELIM@BERT  & 0.16 & 0.27 & 0.22 & $0.27  \pm 0.080$ \\

 \hline
  & OffenseEval@BERT & 0.92 & 0.80 & 0.85 & $0.88 \pm 0.044$  \\
& Dexter@BERT+ & 0.96 & 0.72 & 0.83 & ${0.87} \pm0.048$  \\
& Golden@BERT & 0.84 & 0.87 & 0.85 & $0.87 \pm	0.041$\\

 Cyberbullying  \cite{Cyberbullying} & AbuseEval@BERT  & 0.89 & 0.78 & 0.82 & $0.83  \pm 0.050$\\
& Dexter@BERT & 0.93 & 0.60 & 0.72 & $0.77 \pm 0.051$ \\

& HateEval@BERT  & 0.86 & 0.61& 0.71 & $078 \pm 0.075$ \\

& Dexter-@BERT & 0.91& 0.56 & 0.68 & $0.77 \pm 0.080$ \\
& PRELIM@BERT & 0.80 & 0.57 & 0.67 & $0.70 \pm 0.012$ \\

\hline
\end{tabular}\caption{Results of the various models on the test data sets sorted according to macro F1 score.}
\label{table:ClassificationResults}
\end{table*}

\begin{table*} [ht!]
\small

\label{table:LeakLAMBADA}
\centering
\begin{tabular}{ |c| c | c c c c | }
\hline
\textbf{Dataset}& \textbf{Model} & Precision & Recall &  F1 score &  Macro F1 score   \\
 \hline
 
& { Dexter@BERT+ }& $0.92$ & ${0.87}$ & $ 0.89$ & ${0.91} \pm {0.029}$ \\
& Dexter@BERT & 0.91 & 0.86 & 0.88 & $0.90 \pm 0.037$ \\
Sexual 
Predator Identification & Dexter-@BERT & 0.80 & 0.93 & 0.86 & $0.88 \pm 0.043$\\
Competition \cite{inches2012overview}&LAMBADA@BERT+ & 0.85 & 0.73 & 0.77& $0.80 \pm 0.074$\\

&LeakGAN@BERT+ & 0.55 & 0.99 & 0.71& $0.51 \pm 0.035$ \\
& LeakGAN@BERT  & 0.53 & 0.85 & 0.65 & $0.51 \pm 0.041$ \\
& LAMBADA@BERT & 0.47 & 0.80 & 0.51 & $0.30 \pm 0.019$\\ \hline

&  Dexter@BERT+ & 0.66 & 0.80 & 0.72 & ${0.70} \pm 0.083$ \\
 & Dexter@BERT & 0.55 & 0.88 & 0.67 & $0.54 \pm 0.077$ \\

 Empathy \cite{sharma2020computational}  & Dexter-@BERT & 0.51 & 0.65 & 0.57 & $0.52 \pm 0.075$ \\ 
 &  LAMBADA@BERT+ & 0.40 & 0.67 & 0.50 & $0.33 \pm 0.071$ \\
& LeakGAN@BERT+ & 0.30 & 0.76 & 0.41 & $0.31 \pm 0.122$  \\

& LAMBADA@BERT  & 0.81 & 0.35 & 0.40 & $0.50  \pm 0.145$ \\
& LeakGAN@BERT   & 0.28 & 0.53 & 0.36 & $0.30 \pm 0.040$  \\
 \hline
 & Dexter@BERT+ & 0.96 & 0.72 & 0.83 & ${0.87} \pm0.048$  \\
  & LAMBADA@BERT+ & 0.95& 0.70 & 0.80 & $0.83 \pm 0.060$ \\
 & Dexter@BERT & 0.93 & 0.60 & 0.72 & $0.77 \pm 0.051$ \\
 Cyberbullying \cite{Cyberbullying} & LeakGAN@BERT+  & 0.91 & 0.56 & 0.69 & $0.71  \pm 0.064$\\
 & Dexter-@BERT & 0.91& 0.56 & 0.68 & $0.77 \pm 0.080$ \\
& LAMBADA@BERT  & 0.88 & 0.54& 0.66 & $0.71 \pm 0.078$ \\
 & LeakGAN@BERT  & 0.97 & 0.47 & 0.62 & $0.68 \pm 0.114$  \\
\hline
\end{tabular}\caption{Results of the various models that were trained by different text augmentation techniques, sorted according to macro F1 score.}
\label{table:ClassificationResults2}
\end{table*}

\begin{itemize}
    \item The results for the sexual predators place Dexter@BERT+ and Dexter@BERT at the top two both with respect to the other data augmentation pipelines (Table \ref{table:ClassificationResults2}) and with respect to the abusive speech BERT models (Table \ref{table:ClassificationResults}). In fact,  both LAMBADA@BERT+ and LeakGAN@BERT+ obtained worse results than the baseline OffenseEval@BERT. 
    (F1 score of 0.88 vs 0.8 and lower). 

    \item The results for the empathy dataset in Table \ref{table:ClassificationResults} show that Dexter@BERT+ obtained the highest F1 and macro F1 score.  In Table \ref{table:ClassificationResults2} we see that the performance of Dexter@BERT and its derivatives is far better than the other two pipelines. We also observe a poorer overall performance than the other two datasets, in accordance with the absence of a natural positive class. 

    \item  The results for the cyberbullying dataset in Table \ref{table:ClassificationResults} show that Dexter@BERT+ scored at the top (Macro F1 score 0.87) together with OffenseEval@BERT (0.88) and AbuseEval@BERT (0.87). In Table \ref{table:ClassificationResults2} we again see that Dexter@BERT+ came first, although this time, the gap from LAMBADA@BERT+ is small.

    \item The performance of Dexter@BERT+ is similar to OffenseEval@BERT on the sexual predators and cyberbullying data sets. This is to be expected as these datasets have a clear offensive speech element. The more telling result is the larger gap for the empathy dataset, 0.7 vs 0.64 in Macro F1 score. Indeed lack of empathy has more to do with the psychopathic mind than offensive speech.
    
    \item One can look at our results through the lens of transfer learning, where our dataset was successfully used to facilitate transfer learning from the task of offensive speech detection to the task of predicting various aspects associated with the psychopathic personality.

\end{itemize}

\section{Discussion}
This paper presents a new unsupervised approach for personality data augmentation (PEDANT), trading labeled data with domain expertise. We implement it in a specific pipeline that generates sentences with a psychopathic signature (Dexter). One could ask whether it is feasible to assemble a labeled dataset via platforms such as Amazon Turk. The answer is probably no, as domain expertise is required in the field of personality to correctly label the data. Our work offers a scalable and feasible data augmentation pipeline that circumvents such caveats by taking input from a domain expert in the later stages of the pipeline rather than at the beginning (the data collection step).

The clear conclusion from the evaluation experiments we ran is that our pipeline produced synthetic data with better quality than the other two pipelines (\cite{anaby2020not}, and \cite{LeakGan}) thus highlighting the point that not all data augmentation tasks were born equal. The task of generating synthetic data about flight and travel issues (the examples from the LAMBADA paper) is not the same as generating personality-type text. The same way generating a synthetic dog picture is not the same as generating a CT-scan picture of a brain with a tumor in order to train med students to read such images. 

We expect that our pipeline can be adapted to other domains where high-quality labeled data is lacking and hard to obtain by crowd-sourcing: suicide,  school shooters, etc.



\bibliographystyle{unsrt}  
\bibliography{ijcai22}

\appendix

\section{Sentence completion examples}\label{sec:apdx:completions}
Each example consists of a statement that servers as a prompt for GPT-2, GPT-3, and its completion. Two completions are given, before fine-tuning (GPT-2,GPT-3) and after the fine-tuning on the preliminary dataset (GPT-2). The statements were chosen to entail a natural discriminatory completion between a normal and a psychopathic personality. 
The statement is first, and the completions follow.
{\it

\begin{description}
  \item[\it I take advantage of others whenever I can] \hfill \\ 
  (GPT-2 before) I know how to get out of trouble.\\
   \hfill (GPT-3 davinci) said he, "but they generally take advantage of me. It is the way of the world." "I am glad to hear you say so," said Holmes, laughing; "it is a most comforting reflection.".\\
   \hfill (GPT-3 curie) and I am a very strong negotiator. Yes, I wouldn't say that I am perfect, but I do try to do the right thing by all of my partners.I am not a fan of people who are selfish, narcissistic or narcissistic.\\
    \hfill\\(GPT-2 after) I have a very low empathy for others and I am constantly manipulating people to get what I want.
      
\end{description}

\begin{description}
  \item[\it People are violent]
  \hfill \\ (GPT-2 before) I don't think I was ever taught to be violent.\\
    \hfill (GPT-3 davinci) People are not moral. People are amoral. People are depraved.” The senator was bullish on the prospects of the Republican Party. “The future is ours,” he said, to loud applause. “\\
   \hfill (GPT-3 curie) and stupid, but they're not all violent and stupid, and they're not all crazy. I don't know what the hell happened. But I do know that this is not a good thing. Things have been getting better in this country\\
   \hfill \\ (GPT-2 after) They are the ones who kill animals for fun, rape and torture people, and steal and rob.
\end{description}

\begin{description}
  \item[\it People are selfish] 
  \hfill \\ (GPT-2 before) I don't think we are selfish.\\
      \hfill (GPT-3 davinci) they don’t think about what they’re doing,” said the 17-year-old. “They take the time to stop and think about what they’re doing, maybe this wouldn’t happen\\
   \hfill (GPT-3 curie) It's a part of human nature. But, most importantly, it's a part of the human condition." "This is the fundamental question of all human history," he continued. "Do I believe that people are fundamentally selfish?"\\
   \hfill \\ (GPT-2 after) It's just that, when we see someone in pain, we don't really want to help him.
\end{description}

}

\section{Psychopathic Vector}\label{sec:apdx:vector}

28 words that are significantly collocated with the target word “psychopath” in the
iWeb. These words constitute the “psychopathic vector” used for ranking:
psychopath, sociopath, killer, criminal, murderer, victim, empathy, prison,
monster, liar, rapist, violent, dangerous, sexual, antisocial, sadistic, narcissistic, evil,
insane, manipulative, charming, kill, manipulate, torture, hunt, rape, terrorize, terrify.

\section{Seed Sentences}
\label{sec:apdx:seed}

40 seed sentences for completion by GPT. 20 sentences about self, and 20 about others.

\subsection*{Beliefs about self
}
I take advantage of others whenever I can.
\\
I experience no remorse for harm or injury I have caused to others.
\\
I enjoy manipulating others’ emotions to get what I want.
\\
I am unconcerned with the consequences of my actions.
\\
I feel invulnerable and that nothing can hurt me.
\\
I have no interest in the rights, property, or safety of others.
\\
I feel little empathy to the feelings of others.
\\
I don’t understand others’ needs and feelings.
\\
I act impulsively, without regard for the consequences of my actions.
\\
I am unreliable and irresponsible.
\\
I engage in unlawful or criminal behaviour.
\\
I tend to be angry or hostile toward others.
\\
I tend to get into power struggles with other people.
\\
I gain pleasure and satisfaction by being sadistic or aggressive toward others.
\\
I blame others for my failures or shortcomings and believe my problems are caused by external factors.
\\
I don’t understand my behaviour and motives.
\\
I don’t feel shame or remorse.
\\
I am seeking thrill and excitement.
\\
I usually feel empty and bored.
\\
I feel like a hunter seeking a prey.

\subsubsection*{Beliefs about others}
People are selfish.
\\
Human beings are greedy.
\\
The majority of people are cruel.
\\
The world is full of inconsiderate people
\\
Most people are childish
\\
Most people are arrogant
\\
People I know are irresponsible
\\
People are manipulative
\\
Human beings are deceptive
\\
The majority of people are abusive
\\
The majority of people are dangerous
\\
Most people are exploitative
\\
Most people are untrustworthy
\\
People are violent
\\
People are vulnerable
\\
Human beings are weak
\\
The majority of people are helpless
\\
People are predatory
\\
Most people are an easy prey
\\
The human condition is weak and vulnerable to predation
\end{document}